%% file: iclr2024_conference.tex
\documentclass{article}

\usepackage[preprint]{neurips_2024}

\usepackage[utf8]{inputenc} % allow utf-8 input
\usepackage[T1]{fontenc}    % use 8-bit T1 fonts
\usepackage{url}            % simple URL typesetting
\usepackage{booktabs}       % professional-quality tables
\usepackage{amsfonts}       % blackboard math symbols
\usepackage{amsmath,bm}
\usepackage{nicefrac}       % compact symbols for 1/2, etc.
\usepackage{microtype}      % microtypography
\usepackage{xcolor}         % colors
\usepackage{graphicx}
\usepackage{subcaption}
\usepackage{tikz}
\usetikzlibrary{positioning}
\usepackage{algorithm}
\usepackage{algorithmic}
\usepackage{hyperref}       % hyperlinks
\definecolor{mydarkblue}{rgb}{0,0.08,0.45}
\hypersetup{ % play with the different link colors here
    colorlinks,
    citecolor=mydarkblue,
    filecolor=mydarkblue,
    linkcolor=mydarkblue,
    urlcolor=mydarkblue % set to black to prevent printing blue links
}

\DeclareMathOperator*{\argmin}{argmin}

\input{setting}
\title{Removing Spurious Correlation from\\ Neural Network Interpretations}

\author{%
  Milad Fotouhi, \;Mohammad Taha Bahadori, \;Oluwaseyi Feyisetan,\\ \textbf{Payman Arabshahi}, \;\textbf{David Heckerman}\\
  Amazon and University of Washington, Seattle\\
  \texttt{\{mfotouhi, bahadorm\}@amazon.com}
}

\begin{document}

\maketitle

\begin{abstract}
The existing algorithms for identification of neurons responsible for undesired and harmful behaviors do not consider the effects of confounders such as topic of the conversation. In this work, we show that confounders can create spurious correlations and propose a new causal mediation approach that controls the impact of the topic. In experiments with two large language models, we study the localization hypothesis and show that adjusting for the effect of conversation topic,  toxicity becomes less localized.
\vspace{-0.1in}
\end{abstract}
% \vspace{-0.1in}
% \section{Submission of conference papers to ICLR 2024}
\section{Introduction}
\vspace{-0.1in}
\input{main.tex}

\section{Background}
\input{back}

\section{Methodology}
\input{method}

\section{Preliminary Experiments}
\input{exp}

\section{Discussion and Conclusion}
\input{discuss}
\bibliography{references}
\bibliographystyle{plainnat}

\appendix
\input{app}

\end{document}

%% file: setting.tex
\usepackage{xfrac}
\newcommand{\bfx}{\mathbf{x}}
\newcommand{\bmx}{\bm{x}}

\newcommand{\mry}{\mathrm{y}}

\newcommand{\bma}{\bm{a}}
\newcommand{\mra}{\mathrm{a}}

\newcommand{\bfz}{\mathbf{z}}
\newcommand{\bmz}{\bm{z}}

\newcommand{\bfn}{\mathbf{n}}
\newcommand{\bmn}{\bm{n}}

\newcommand{\mrm}{\mathrm{m}}

\newcommand{\bfq}{\mathbf{q}}
\newcommand{\bmq}{\bm{q}}

\newcommand{\bmg}{\bm{g}}

\newcommand{\indep}{\perp \!\!\! \perp}

%% file: main.tex
One of the central approaches to improve safety of Large Language Models (LLMs) is to identify and neutralize units in the LLMs that are responsible for undesired behaviors such as bias and toxicity. Interpretability approaches, such as Mechanistic Interpretability \citep{Olah2020-vi,olsson2022context,nanda2023progress,bereska2024mechanistic}, try to accomplish this by finding subgraphs of the LLM's computational graph that are responsible for the undesired behaviors. Existing work have shown the success of neuron neutralization approach in removing bias and toxicity from the output of LLMs \citep{vig2020investigating,yu2023unlearning,zhang2024towards}.

The first step in editing LLMs is to identify the computational units responsible for harmful behavior. The existing algorithms such as activation patching have focused on identifying mechanisms that are predictive of an undesired behavior \citep{zhang2024towards}. However, confounders such as the conversation topic can create spurious correlations. For example, some topics such as politics are more susceptible to toxic language and we should identify nodes that are activated during toxic generations, not all generations on the politics topic. In this work, we will focus on removing the effect of topic and its spurious effect and identify computational units that are only responsible for the unwanted behavior.

% Suppose we have a large neural network $M$ with known weights and a Q\&A dataset $S$. We run the LLM on the questions and label the answers as toxic or non-toxic. Given the randomness in LLM's generations, we run LLM with different inference hyper-parameters such as temperature and output length. 

%change in the outputs between $S_1$ and $S_2$.
Our task is to find a small subset of neurons that are responsible for toxicity in the LLM's answers. 
We follow the causal mediation analysis approach  \citep{vig2020investigating,meng2022locating,stolfo2023mechanistic,marks2024sparse,zhang2024towards}. In our setup, the treatment (question) takes the form of text and the outcome is the toxicity label of the generated response. To adjust for the effects of topic, we use the weighting approach by \citep{huber_direct_2020, huang2024nonparametric}, which requires only one forward pass. We customize the entropy balancing approach \citep{huang2024nonparametric} to be able to handle textual queries using sentence embeddings. We also propose an approach for finding the average indirect effect without creating out-of-distribution examples. 

We use our proposed approach to study the localization hypothesis that states only few neurons are responsible for harmful behavior in LLMs. We use two LLMs Bloomz 1B7 \citep{le2023bloom} and Microsoft/Phi-3-mini 3B \citep{abdin2024phi} on the RealToxicityPrompts dataset \citep{gehman2020realtoxicityprompts}. We compute how much each multilayer perceptron unit is responsible for toxicity. We show that adjusting for the conversation topic, the toxicity become more distributed across MLPs, for both Bloomz and Phi-3-mini.

%% file: back.tex
We use Roman letters for random variables, e.g., $\mathrm{x}$, and italic letters for scalar values $x$. Random vectors are denoted by bold Roman letters $\mathbf{x}$ and vector of values are denoted by italic bold letters $\bm{x}$.

\subsection{Backround on Robust Causal Inference}
To describe the language of potential outcomes \citep{imbens2015causal,chernozhukov2024applied}, used in this paper, consider the classical problem of binary treatment effect estimation. We have a binary treatment $\mathrm{a} \in \{0, 1\}$ and outcome $\mathrm{y} \in \mathbb{R}$. Both treatment and outcome are influenced by a $d$-dimensional confounder (pre-treatment covariates) $\mathrm{x} \in \mathbb{R}^{d}$. 
Define the potential outcome $\mathrm{y}{\{\mathrm{a}=a\}}$ as the outcome when we intervene and set the value of the treatment to $a$. In real world, we only observe one of the potential outcomes for either $\mathrm{a}=0$ or $\mathrm{a}=1$.
The average treatment effect is defined as $\mathbb{E}\left[\mathrm{y}{\{1\}}-\mathrm{y}{\{0\}}\right]$.

Under classical conditions of ``strong ignorability'', $\mry{\{\mra\}} \indep \mra~|~\bfx$, (i.e., no hidden confounders) and ``positivity'', $0 < \mathbb{P}(\mra | \bfx) < 1$, there are two primary approaches to estimate the average treatment effect \textit{Outcome Regression} and \textit{Inverse Propensity-score Weighting} (IPW). In the outcome regression approach, we train a regression model for the outcome $\mry = f(\mra, \bfx)$. We use the regression model to compute the potential outcomes for both $\mra = 0, 1$ and estimate the average treatment effect by finding the empirical average of $\mathrm{y}{\{1\}}-\mathrm{y}{\{0\}}$. 

In the IPW approach, the propensity scores are defined as $\pi(\bfx)= \mathbb{P}[\mra=1|\bfx]$. Weighting each data point by the inverse propensity scores, creates a pseudo-population in which the confounders and treatments are independent. Thus, regular regression algorithms can estimate the causal response curve using the pseudo-population, which resembles data from randomized trials. When the treatments are continuous, i.e., $\mra \in \mathbb{R}$, we commonly use the stablized weights $\pi = \sfrac{f(a)}{f(a|\bmx)}$ \citep{robins2000marginal,zhu2015boosting}, where $f$ denotes the marginal density function. We can also cleverly combine the outcome regression and IPW to achieve robustness to misspecifications in the regression or propensity score models \citep{bang2005doubly,chernozhukov2018double}.

The main challenge in estimating average treatment effect with IPW is that the weights can be very large for some of the data points, leading to unstable estimations. \citet{kang2007demystifying} and \citet{smith2005does} provide multiple pieces of evidence that the propensity score methods can lead to large biases in the estimations. To address the problem of extreme weights, \textit{Entropy Balancing (EB)} \citep{hainmueller2012entropy} estimates weights such that they balance the confounders subject to a measure of dispersion on the weights to prevent extremely large weights. Here we describe EB for multi-dimensional continuous treatments based on \citep{bahadori2022end}. We first center both the treatments $\bma$ and the confounders $\bmx$ vectors. We create a vector $\bmg(\bma, \bmx) = [\bma, \bmx, \bma\otimes\bmx]$, where $\otimes$ denotes the outer product. Entropy balancing maximizes the entropy of the propensity scores:
\begin{align*}
    \widehat{\bm{\pi}}= \argmin_{\bm{\pi}}~\bm{\pi}^{\top}\log\left(\bm{\pi}\right),\label{eq:primal}
    \quad\text{s.t.   } \quad
    \text{(i)}~  \bm{G}\bm{\pi} = \bm{0}, \quad
    \text{(ii)}~ \bm{1}^{\top}\bm{\pi} = 1, \quad
    \text{(iii)}~ \bm{\pi} \succeq 0,
\end{align*}
where matrix $\bm{G}$ denotes the concatenation of $\bmg$ for all data points. We often solve the dual of the above problem, by the following optimization (see the derivation in \citep{wang2020minimal}):
\begin{align*}
    \widehat{\bm{\lambda}} &= \argmin_{\bm{\lambda}}\; \log\left(\bm{1}^{\top}\exp\left(- \bm{\lambda}^{\top}\bm{G}\right) \right)+\gamma \|\bm{\lambda}\|_1.
\end{align*}
We have added the regularization for approximate balancing \citep{wang2020minimal}. Higher values of the regularization penalty pushes the weights closer to the confounded (equal weight) solution. Given a solution $\widehat{\bm{\lambda}}$, the weights are obtained as $\bm{\pi} = \mathrm{softmax}\left(- \widehat{\bm{\lambda}}^{\top}\bm{G}\right)$. 

\subsection{Causal Mediation Analysis for Neural Network Interpretation}
In contrast to estimation of the effect of a cause, reviewed in previous section, causal mediation analysis answers the questions about \textit{Causes of Effect}  \citep{dawid2000causal,dawid2014fitting,lu2023evaluating,kawakami2024probabilities,zhao2023conditional}. In the simplest mediation analysis setup, we assume that we have a binary treatment $\mra$ whose effect has two paths to reach to the outcome $\mry$: the direct path and an indirect path that goes through a mediation variable $\mrm$. To quantify the effect that goes from each path, we need to use nested counterfactuals. The counterfactual $\mry{\{a, \mrm{\{a'\}}\}}$ is interpreted as the outcome when we intervene and set the treatment to $a$ but the input to the mediator to $a'$. 

Using the notion of nested counterfactual, \citet[\S 4.5]{pearl2009causality} defines the relationship between the treatment effect and mediation quantities:
\begin{align*}
   &\underbrace{\mry{\{a=1, \mrm{\{a'=1\}}\}} - \mry{\{a=0, \mrm{\{a'=0\}}\}}}_{\text{Treatment Effect}}\\
   &= 
   \underbrace{\mry{\{a=1, \mrm{\{a'=1\}}\}} - \mry{\{a=0, \mrm{\{a'=1\}}\}}}_{\text{Direct Effect}} + 
   \underbrace{\mry{\{a=0, \mrm{\{a'=1\}}\}} - \mry{\{a=0, \mrm{\{a'=0\}}\}}}_{\text{Indirect Effect}}
\end{align*}
Common approaches for mediation analysis fit a model for the outcome in terms of the treatment, mediator, and confounders and estimate the quantities above \citep{imai2010general,rijnhart2021mediation}.

Existing methods for neural networks interpretation \citep{vig2020investigating,meng2022locating,zhang2024towards} implement interventions by corrupting the token embeddings in the input prompt. 
In particular, The activation patching (also known as causal tracing) 
 performs three forward runs of the prompt: (1) clean prompt, (2) corrupted prompt, and (3) corrupted prompt where the target token is replaced by the uncorrupted token. However, this approach does not remove the impact of confouders' effect on the relationships they find. In the next section, we propose a method for removing the spurious effects from neural network interpretations.

%% file: method.tex
\begin{figure}[t]
    \centering
\begin{tikzpicture}[node distance={15mm}, thick, main/.style = {draw, circle, line width=1.5pt}]
    % Nodes
    \node[main] (X) [align=center, draw=red, text=red]  (X) {\(\bfx\)};
    \node[main] (Q) [below left=of X, align=center] (Q) {\(\bfq\)};
    \node[main] (Y) [below right=of X, align=center] (Y) {\(\mry\)};
    \node[main] (N) [below=23mm of X, align=center] (N) {\(\bfn\)};
    
    % Edges
    \draw[->, red, line width=1.5pt] (X) -- (Q);
    \draw[->, red, line width=1.5pt] (X) -- (Y);
    \draw[->, line width=1.5pt] (Q) -- (Y);
    \draw[->, line width=1.5pt] (Q) -- (N);
    \draw[->, line width=1.5pt] (N) -- (Y);
\end{tikzpicture}
    \caption{We quantify the Natural Indirect Effect (NIE) in conversations using the above DAG. The conversation topic $\bfx$ influences both the Question $\bfq$ and the harmfulness (e.g., bias or toxicity) of the LLM generations $\mry$. Our goal is to use this graph for every node and find the amount of effect that goes through a particular internal node $\bfn$ (e.g., activations of internal neurons). Previous studies do not consider the impact of conversation topic $\bfx$ (marked by red color). }
    \label{fig:single-node}
\end{figure}
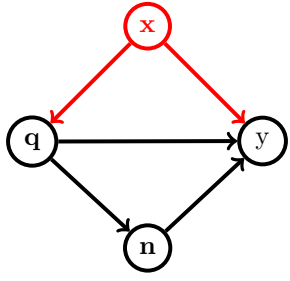

To quantify the contributions of a computational unit of an LLM (such as an MLP) in toxicity of its output, we propose the directed acyclic graph in Figure \ref{fig:single-node}. In our problem setting, the treatments are the queries (prompts) $\bfq$, the mediators are the intermediate node activations $\bfn$, the confounders are the conversation topic $\bfx$, and the outcome is the harmfulness label $\mry$. We use bold face for the queries to emphasize that we use a vector representation of the input query. Our goal is to use this graph for every node and find the amount of effect that goes through it.

Similar to the binary case, we can do the mediation analysis with multivariate treatment as follows.
Define the response function as $\mu(\bmq, \bmq') = \mathbb{E}[\mry{\{\bmq, \bfn\{\bmq'\}\}}]$. We can decompose the average treatment effect when the treatment changes from $\bmq$ to $\bmq'$ as follows:
\begin{equation*}
\underbrace{\mu(\bmq', \bmq') - \mu(\bmq, \bmq)}_{\text{ATE}} = \underbrace{\mu(\bmq', \bmq') - \mu(\bmq, \bmq')}_{\text{NDE}} + \underbrace{\mu(\bmq, \bmq') - \mu(\bmq, \bmq)}_{\text{NIE}},
\end{equation*}
where the terms on the right hand side Natural Indirect Effect (NIE) and Natural Direct Effect (NDE) measure the amount of effect going through the mediating node and all other nodes, respectively. Thus, NIE is our the quantity of interest.

Mediation analysis for non-binary treatments is not extensively studied. For general treatments, \citet{huber_direct_2020} and \citet{huang2024nonparametric} provide the following equation for estimating the NIE:
\begin{align}
    \mu(\bmq, \bmq') &= \mathbb{E}_{\bfn, \bfx}\left[\frac{\pi_{\bfn, \bfx}(\bfq, \bfn, \bfx)}{\pi_{\bfn, \bfx}(\bfq+\bm{\delta}, \bfn, \bfx)}\pi_{\bfx}(\bfq+\bm{\delta}, \bfx)\mry \left| \bfq=\bmq\right. \right], \label{eq:estimate}\\
    \pi_{\bfz}(\bfq, \bfz) &= \frac{f_\bfq(\bfq)}{f_{\bfq|\bfz}(\bfq|\bfz)}, \qquad
    \bm{\delta} = \bm{q}' - \bm{q}. \nonumber
\end{align}
In this equation, we use the stablized form of the weights in $\pi_{\bfz}(\bfq, \bfz)$ as in \citep{robins2000marginal}. Parsing the equation, note that for $\bm{\delta}=\bm{0}$, Eq. (\ref{eq:estimate}) reduces to average treatment effect: $\mu(\bmq, \bmq) = \mathbb{E}[\pi_\bfx(\bmq, \bfx)\mry]$. When there is no confounding, we have $\mu(\bmq, \bmq') =\mathbb{E}_{\bfn}\left[\frac{\pi_{\bfn}(\bfq, \bfn)}{\pi_{\bfn}(\bfq+\bm{\delta}, \bfn)}\mry \left| \bfq=\bmq\right. \right]$. This equation means the mediation increases if the query $\bmq'$ is likelier than $\bmq$ in the propensity score model. 
Finally, note that the approach in Eq. (\ref{eq:estimate}) does not require additional training. It only requires a single forward pass.
%If we extract the topic $X$ from queries, we can simplify $\pi_{N, X}(Q, N, X) = \pi_{N}(Q, N)$

Unlike \citep{huber_direct_2020, huang2024nonparametric}, who are interested in estimating $\mu(\bmq, \bmq')$, our target quantity is the \textit{Average Indirect Effect} (AIE). In the following estimator, we measure the amount of NIE for each pairs of the queries in the data to avoid out-of-distribution data:
\begin{align}
\text{AIE}  &= \frac{1}{m(m-1)} \sum_{\substack{i,j=1\\i\neq j}}^{m}\mu(\bmq_i, \bmq_j)-\mu(\bmq_i, \bmq_i) \nonumber\\ 
&= \frac{1}{m(m-1)}\sum_{\substack{i,j=1\\i\neq j}}^{m}\left\{\frac{\pi_{\bfn, \bfx}(\bmq_i, \bmn_i, \bmx_i)}{\pi_{\bfn, \bfx}(\bmq_j, \bmn_i, \bmx_i)}\pi_{\bfx}(\bmq_j, \bmx_i)-\pi_{\bfx}(\bmq_i, \bmx_i)\right\}y_i. \label{eq:emp}
\end{align}
For binary outcomes $y$, the loop can be computed in linear time $O(m\cdot m_{nz})$, where $m_{nz}$ denotes the number of non-zero outcomes. In practice, we implement a stochastic approximation of the sum by computing only 200 $\bmq_j$ for every $\bmq_i$ (where $y_i=1$) to further improve speed. For robustness, we find Winsorized mean by clamping the top and bottom 5-percentile of the quantities inside the summation \citep{cole2008constructing,crump2009dealing,chernozhukov2018double}.

\paragraph{Entropy Balancing.} Simple attempts at computing the propensity scores in Eq. (\ref{eq:emp}) with high dimensional variables, such as the parametric approach in Appendix \ref{sec:param}, will lead to extreme propensity scores and unstable AIE values. Thus, we follow \citet{huang2024nonparametric} to propose a non-parametric estimator using entropy balancing. While entropy balancing provides an estimate for $\pi_\bfz(\bmq_i, \bmz_i)$, it does not naturally provide an estimate for $\pi_\bfz(\bmq_j, \bmz_i)$, because of the normalization in softmax. To resolve this issue, we rewrite the term inside the sum in Eq. (\ref{eq:estimate}) as follows:
\footnotesize
\begin{align*}
    \frac{\pi_{\bfn, \bfx}(\bmq_i, \bmn_i, \bmx_i)}{\pi_{\bfn, \bfx}(\bmq_j, \bmn_i, \bmx_i)}\pi_{\bfx}(\bmq_j, \bmx_i)-\pi_{\bfx}(\bmq_i, \bmx_i) &= \pi_{\bfx}(\bmq_i, \bmx_i) \left(\frac{\pi_{\bfn, \bfx}(\bmq_i, \bmn_i, \bmx_i)}{\pi_{\bfn, \bfx}(\bmq_j, \bmn_i, \bmx_i)}\frac{\pi_{\bfx}(\bmq_j, \bmx_i)}{\pi_{\bfx}(\bmq_i, \bmx_i)} -1\right)\\
    &=\pi_{\bfx}(\bmq_i, \bmx_i)\left(\frac{\exp-\bm{\lambda}_{\bfn,\bfx}^{\top}\bm{g}(\bm{q}_i, [\bm{n}_i, \bm{x}_i])}{\exp-\bm{\lambda}_{\bfn,\bfx}^{\top}\bm{g}(\bm{q}_j, [\bm{n}_i, \bm{x}_i])}\frac{\exp-\bm{\lambda}_{\bfx}^{\top}\bm{g}(\bm{q}_j, \bm{x}_i)}{\exp-\bm{\lambda}_{\bfx}^{\top}\bm{g}(\bm{q}_i, \bm{x}_i)} -1\right),
\end{align*}
\normalsize
where $\bm{\lambda}_\bfx$ and $\bm{\lambda}_{\bfn, \bfx}$ are the solutions of the entropy balancing with $\bfz$ variable being $\bfx$ and $[\bfn, \bfx]$, respectively. The above result is because the softmax normalization terms cancel out and we can compute the ratios. We provide the full details in Algorithm \ref{alg:ebsolution}.

%An alternative approach for robust computation of the propensity score ratios is to use classification algorithms, see \citep[ pt. II, ch. 4]{sugiyama2012density} and \citep{arbour2021permutation,quintas2024multiply}. We present the full details in Algorithm \ref{alg:ebsolution}. 

\begin{algorithm}[tb]
  \caption{AIE Estimation via Entropy Balancing for a specific unit}
  \label{alg:ebsolution}
\begin{algorithmic}[1]
\REQUIRE $\{\bm{q}_i, \bm{n}_i, \bm{x}_i, y_i\}_{i=1}^{m}$: Question embedding, node activations, topic vectors, and harmfulness outcomes for $m$ questions. Constant $K$.
\STATE $\mathcal{I}_{nz} \gets \{i| y_i=1\}$
\STATE $\bm{\lambda}_{\bfx} \gets EB(\{\bm{q}_i, \bm{x}_i\}_{i=1}^{m})$.
\STATE $\bm{\pi}_\bfx \gets \mathrm{softmax}(-\bm{\lambda}_\bfx^{\top}\bm{G}(\bmq, \bmx))$.
\STATE $\bm{\lambda}_{\bfx, \bfn} \gets EB(\{\bm{q}_i, \bm{n}_i, \bm{x}_i\}_{i=1}^{m})$.
\STATE $AIE \gets \emptyset$.
\FOR{$i \in \mathcal{I}_{nz}$}
\STATE $\mathcal{I}_j \gets$ $K$ randomly selected queries.
\FOR{$j \in \mathcal{I}_j$}
\STATE $AIE \gets AIE \cup \left\{ \frac{\exp-\bm{\lambda}_{\bfn,\bfx}^{\top}\bm{g}(\bm{q}_i, [\bm{n}_i, \bm{x}_i])}{\exp-\bm{\lambda}_{\bfn,\bfx}^{\top}\bm{g}(\bm{q}_j, [\bm{n}_i, \bm{x}_i])}\pi_{\bfx}(\bmq_i, \bmx_i)\left(\frac{\exp-\bm{\lambda}_{\bfx}^{\top}\bm{g}(\bm{q}_j, \bm{x}_i)}{\exp-\bm{\lambda}_{\bfx}^{\top}\bm{g}(\bm{q}_i, \bm{x}_i)} -1\right)\right\}$ .
\ENDFOR
\ENDFOR
\STATE \textbf{Return} WinsorizedMean($AIE$).
\end{algorithmic}
\end{algorithm}

% \vspace{-0.05in}
\paragraph{A Note on Positivity Assumption.} Per \citet{huber_direct_2020}, valid inference with Eq. (\ref{eq:estimate}) requires $f(\bmq|\bmn, \bmx) > 0$ for any $\bmq$, $\bmn$, and $\bmx$ in their valid domain. Because of the deterministic relationship between node activations and questions, this assumption might be violated and consequently the propensity scores might become extreme. We alleviate this by reducing the dimensionality of questions and activations, which increases the likelihood of activations generated by other questions.

%% file: exp.tex
We use the RealToxicityPrompts dataset \citep{gehman2020realtoxicityprompts} for our studies. We ask the LLMs to complete the prompts. We use GPT4 to label the responses as toxic or non-toxic. Both of our instruction-tuned LLMs (Bloomz and Phi-3-mini) are tuned to not respond with toxic language. Thus, only a small fraction of results are labeled as toxic: 7.5\% and 5.2\% for Bloomz and Phi-3-mini, respectively.

We obtain the query representation using RoBERTa \citep{roberta2019} with embedding dimension 768. %Figure \ref{fig:tnse} shows the t-SNE embedding of the query representations. 
We use k-means to cluster the queries into 3 cluster of topics, shown in Figure \ref{fig:tnse} in the appendix. We verify that clusters have different rates of toxicity levels: [8.7\%, 6.6\%, 7.5\%] and [5.9\%, 4.8\%, 5.0\%] for Bloomz and Phi-3-mini, respectively.

We record the activations of the MLPs during generations of the outputs. Bloomz has 25 (1536 dimensional) and Phi-3-min has 33 (3072 dimensional) MLPs. Given the large dimensionality of $\bm{q}\otimes \bm{n}$ in the entropy balancing approach, we use PCA to reduce the dimensionality of the hidden states and embeddings to 25.

\begin{figure}[htbp]
    \centering
    \begin{subfigure}[b]{0.49\textwidth}
        \centering
        \includegraphics[width=\textwidth]{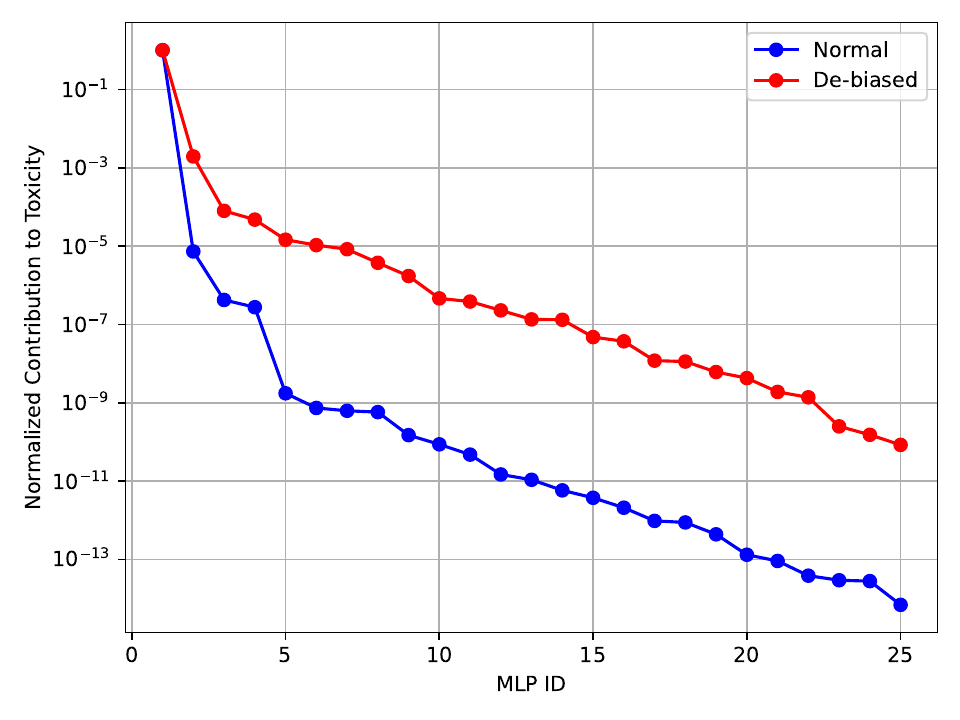}
        \caption{Bloomz}
        \label{fig:bloomz}
    \end{subfigure}
    \hfill
    \begin{subfigure}[b]{0.49\textwidth}
        \centering
        \includegraphics[width=\textwidth]{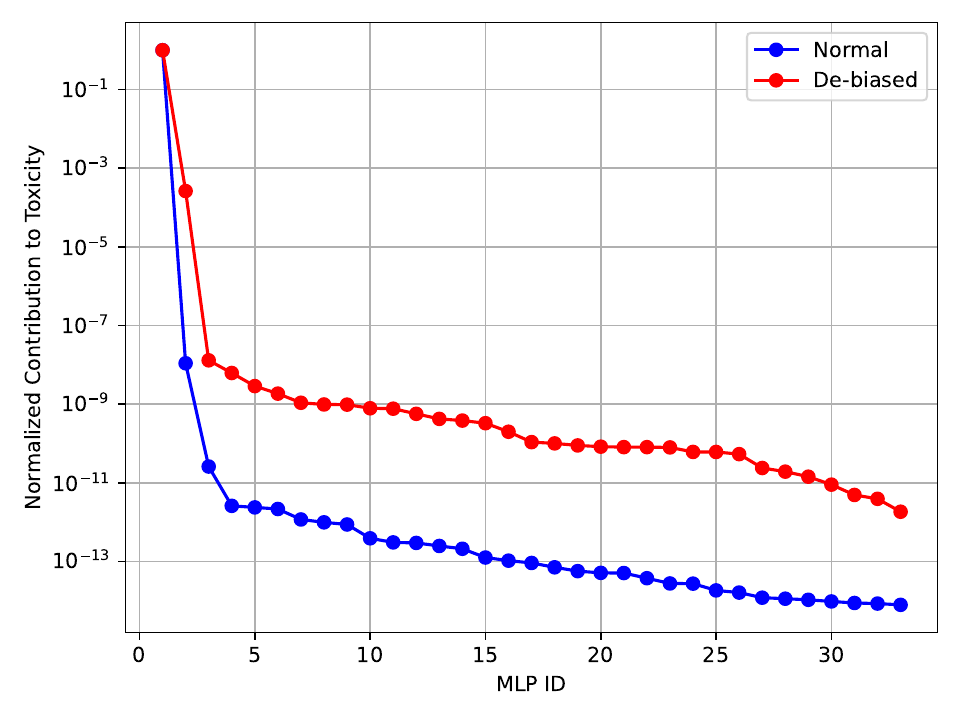}
        \caption{Phi-3-mini}
        \label{fig:phi3}
    \end{subfigure}
    \caption{Contributions of different MLPs in generation of toxic outputs. We measure the average indirect effect, mediated by each MLP.}
    \label{fig:Contributions}
\end{figure}

Figures \ref{fig:bloomz} and \ref{fig:phi3} show the average indirect effects for different MLPs for Bloomz and Phi-3-mini, respectively. The baseline is labeld as `Normal', which is obtained by Eq. (\ref{eq:estimate}) without confounding factors. In both figures, we see that removing the spurious effects of the topics, the slope of the curves, especially for Bloomz, decrease. This means that more MLPs are responsible for toxic outputs, implying a lower degree of toxicity localization as we discussed in the introduction.

%% file: discuss.tex
In this work, we proposed a method for adjusting for the effects of discussion topic on identification of the units that are responsible for the undesired behavior. Our empirical study shows that such adjustment can significantly change the attribution of the behavior to the computational units. As future work, we plan to mask the identified nodes and show that the toxicity decreases. We can also use approaches other than clustering for defining the conversation topics.

The DAG in Figure \ref{fig:single-node} is quite general and can be used for de-biasing the effects of any spurious correlations. For example, we expect gender bias, hallucination, and factuality to be confounded by the conversation topic as well. We also expect removing the confounding bias will impact mechanistic interpretability results on these tasks.

% One of the key goals of mechanistic interpretability is to identify the set of neurons (within an MLP) that work together \citep{turner2023activation,gurnee2024language,park2024linear}. In our proposed formulation, we can identify such subspace as a direction $\nu$ that maximally increase the level of mediation:
% \begin{align}
% \bm{\nu}^{\star} & = \nabla_{\bm{\nu}} \left(\frac{\pi_{N, X}(\bm{q}_i, \bm{n}_i+\bm{\nu}, \bm{x}_i)}{\pi_{N, X}(\bm{q}_j, \bm{n}_i+\bm{\nu}, \bm{x}_i)}\pi_{X}(\bm{q}_j, \bm{x}_j)y_i \right) \nonumber\\
% & = \sum_{\substack{i,j=1\\i\neq j}}^{m}\pi_{X}(\bm{q}_j, \bm{x}_j)y_i\nabla_{\bm{\nu}} \left(\frac{\pi_{N, X}(\bm{q}_i, \bm{n}_i+\bm{\nu}, \bm{x}_i)}{\pi_{N, X}(\bm{q}_j, \bm{n}_i+\bm{\nu}, \bm{x}_i)} \right)\nonumber\\
% &=\sum_{\substack{i,j=1\\i\neq j}}^{m}\pi_{X}(\bm{q}_j, \bm{x}_j)y_i\nabla_{\bm{\nu}} \left(\frac{\exp\left\{-\bm{\lambda}^{\top} \bm{g}(\bm{q}_i, [\bm{n}_i+\bm{\nu}, \bm{x}_i])\right\}}{\exp\left\{-\bm{\lambda}^{\top} \bm{g}(\bm{q}_j, [\bm{n}_i+\bm{\nu}, \bm{x}_i])\right\}} \right), \nonumber
% \end{align}
% where the last step is due to the form of the entropy balancing solution for the propensity scores. This gradient can be further simplified. Using this approach, we can find the direction for each MLP that has the largest gradient.

%% file: app.tex
\section{Parametric Estimation of Propensity Scores} 
\label{sec:param}
We need to have estimates for two conditional distributions: $f(q|X)$ and $f(q|N, X)$. Because $X$ is categorical, the first one is simply computed as $|X|$ number of multivariate Gaussian distributions with different mean vector and (diagonal) covariance matrix $f(q|x_k) = \mathrm{MVN}(\bm{m}_k, \Sigma_k)$. We use a linear regression model to estimate $f(q|N, X)$ as $\mathbb{E}[Q|N,X] = W[N, X]$. We approximate the residual using a  multivariate Gaussian distributions with zero mean and (diagonal) covariance matrix. Thus, the distribution becomes: $f(q|N, X) = \mathrm{MVN}(W[N, X], \widetilde{\Sigma})$. Given the marginal distribution $f(q) = \mathrm{MVN}(\bm{m}_q, \Sigma_q)$, we can simplify:
\begin{align*}
    &\frac{\pi_{N, X}(q, n, x)}{\pi_{N, X}(q', x)}\pi_{X}(q', x) = \frac{f(q)f(q'|n, x)}{f(q|n,x)f(q'|x)}\\
    &= \sqrt{\frac{|\Sigma_k|}{|\Sigma_q|}}\exp\left\{\mathrm{Qd}(\bm{q}, \bm{m}_q, \Sigma_q)+\mathrm{Qd}(\bm{q}', W[\bm{n}, \bm{x}], \widetilde{\Sigma})-\mathrm{Qd}(\bm{q}', \bm{m}_k, \Sigma_k)-\mathrm{Qd}(\bm{q}, W[\bm{n}, \bm{x}], \widetilde{\Sigma})\right\},
\end{align*}
where the quadratic function is defined as $\mathrm{Qd}(\bm{a}, \bm{m}, \Sigma) = -1/2\cdot(\bm{a}-\bm{m})^{\top}\Sigma^{-1}(\bm{a}-\bm{m})$.

\begin{figure}
    \centering
    \includegraphics[width=0.5\linewidth]{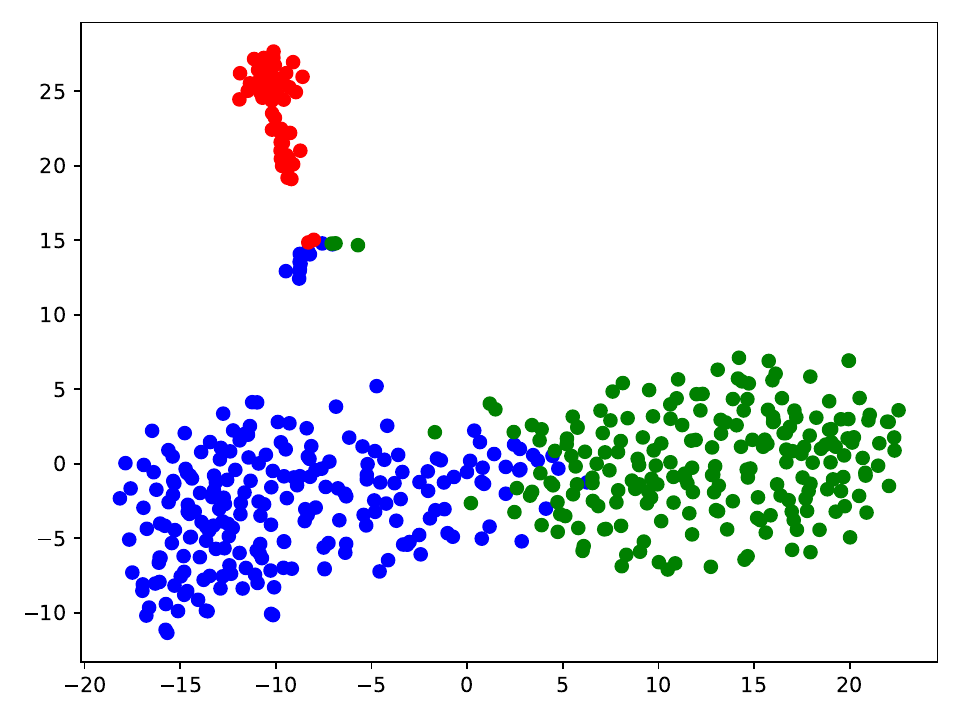}
    \caption{The t-SNE embedding of 1000 randomly selected queries and the clusters identified by k-means in the RealToxicityPrompts dataset \citep{gehman2020realtoxicityprompts}.}
    \label{fig:tnse}
\end{figure}